\documentclass[10pt, a4paper]{article}

\usepackage[final]{lrec2026} 
\usepackage{natbib}
\usepackage{enumitem}
\usepackage{booktabs} 
\usepackage{multirow} 
\usepackage{xcolor}
\usepackage{makecell}
\usepackage[linesnumbered,ruled,vlined]{algorithm2e}
\usepackage{amsmath, amssymb}

\usepackage{pifont}
\usepackage{caption}
\usepackage{subcaption}
\newcommand{\fullcirc}{\ding{108}} 
\newcommand{\emptcirc}{\ding{109}} 

\newcommand{\circsep}{\hspace{0.35em}}
\newcommand{\F}{\fullcirc\circsep}
\newcommand{\E}{\emptcirc\circsep}

\newcommand{\hetscale}[1]{%
\ifcase#1
\E\E\E\E\E\emptcirc\or
\F\E\E\E\E\emptcirc\or
\F\F\E\E\E\emptcirc\or
\F\F\F\E\E\emptcirc\or
\F\F\F\F\E\emptcirc\or
\F\F\F\F\F\emptcirc\or
\F\F\F\F\F\fullcirc
\fi
}

\title{Optimizing Multilingual LLMs via Federated Learning: \\ A Study of Client Language Composition}

\name{Aleix Sant$^1$, Jordi Luque$^1$, Carlos Escolano$^2$ \vspace{5pt}} 
\address{Scientific Research, Telefónica Innovación Digital \vspace{1pt} \\
        Universitat Politècnica de Catalunya \vspace{1pt} \\
        Barcelona, Spain \vspace{5pt} \\
         \{aleix.santsavall, jordi.luqueserrano\}@telefonica.com \vspace{1pt} \\ 
         carlos.escolano@upc.edu\\}

\abstract{
Federated Learning (FL) of Large Language Models (LLMs) in multilingual environments presents significant challenges stemming from heterogeneous language distributions across clients and disparities in language resource availability. To address these challenges, we extended the \textit{FederatedScope}-LLM framework to support multilingual instruction-tuning experiments with LLMs. We also introduced a novel client-specific early stopping mechanism, Local Dynamic Early Stopping (LDES-FL), which allows clients to pause and resume local training based on client-side validation performance, enhancing training efficiency and sustainability. Through a series of experiments, we studied how client language composition — from fully monolingual to increasingly multilingual clients — affects multilingual quality, fairness and training cost. Monolingual local fine-tuning remains the most effective for single-language specialization, whereas federated training is better suited to learning a single balanced multilingual model. In FL, increasing within-client multilinguality leads to stronger and fairer global models, narrows the gap to centralized multilingual fine-tuning, and yields the largest gains for lower-resource languages, albeit at the cost of more optimization steps. Overall, our results identify client language composition as a key design variable in multilingual FL, shaping performance, fairness and efficiency.
\\ \newline }

\begin{document}

\maketitleabstract

\section{Introduction}
\label{sec:intro}
Federated Learning (FL) is a distributed machine learning paradigm that enables collaborative model training across multiple clients while preserving data privacy. Unlike traditional centralized approaches, FL keeps data localized: clients train models on their private data and share only model updates (e.g., weights or gradients) with a central server (the aggregator). This design reduces privacy risks and supports compliance with regulations such as GDPR. The standard FL workflow involves a server distributing a global model, clients performing local training, and the server aggregating their updates. 

Despite its potential, applying FL to Large Language Models (LLMs) \cite{10759678,10733964} introduces challenges, particularly in communication and computation for resource-constrained clients. Parameter-Efficient Fine-Tuning (PEFT) methods, such as Low-Rank Adaptation (LoRA) \cite{DBLP:conf/iclr/HuSWALWWC22}, mitigate these issues by reducing trainable parameters and communication overhead \cite{DBLP:conf/nips/Ye0ZCDL0C24}. Another major challenge is data heterogeneity. In real-world FL deployments, client data is often non-IID (i.e., not Independent and Identically Distributed), which impairs convergence and degrades generalization performance for classical algorithms like FedAvg \cite{pmlr-v54-mcmahan17a}. This phenomenon, commonly referred to as \textit{client drift} \cite{10468591}, also arises in multilingual FL scenarios \cite{DBLP:conf/naacl/WellerMBLD22, 10.1145/3485447.3511988}, where each client may hold data in a different language, representing a specific type of non-IID distribution. Such linguistic diversity induces highly skewed distributions \cite{DBLP:conf/chil/ManoelGBSCSMKD23}, complicating the optimization process and slowing convergence, as local updates tend to diverge from the global objective.

In this work, we investigate how the distribution of multilingual and monolingual data in clients affects performance, fairness and convergence behavior in the federated fine-tuning of multilingual LLMs. To this end, we design a series of experimental scenarios in which the language composition of each client’s local dataset is systematically varied. These scenarios range from fully monolingual settings -- where each client contains data in a single language -- to fully multilingual ones, where each client holds data from several languages, thereby approximating increasingly IID data distributions within the FL framework. 
Our experiments show a common FL pattern: stronger cross-client heterogeneity (here, a larger share of disjoint monolingual data across clients) increases client drift and harms the global solution. Making clients more multilingual improves average multilingual performance and reduces cross-lingual disparities (i.e., increases multilingual fairness) -- especially for lower-resource languages -- at the cost of additional optimization steps.

To carry out this research, we built upon the \textit{FederatedScope}-LLM framework \cite{DBLP:conf/kdd/KuangQLCGPXLDZ24}, extending its capabilities to obtain a flexible repository\footnote{\label{fn:repo}\scriptsize\url{https://github.com/Telefonica-Scientific-Research/FedEloquence}}, designed for multilingual FL experiments with LLMs. The main contributions of this paper are listed below, with the corresponding implementation provided in the repository:

\begin{enumerate}
  \item \textbf{Multilingual FL support:}  
        We add explicit multilingual support for federated fine-tuning of LLMs with flexible prompt integration, language-aware sample processing and multilingual FL data pipelines.
  \item \textbf{Local Dynamic Early Stopping (LDES-FL):}  
        A client-level early stopping mechanism based on local validation loss, allowing clients to pause and resume training dynamically, enabling a more nuanced analysis of multilingual FL dynamics and reducing unnecessary computation.
  \item \textbf{Federated FT experiments on different language client compositions:} 
        We fine-tuned \texttt{salamandra-2b-instruct} across multiple FL settings with varying degrees of within-client multilinguality, ranging from fully monolingual to highly multilingual clients.
\end{enumerate}

\section{Related Work}
\label{sec:related}
Federated Learning is an active area of privacy-aware research that has been extensively studied across domains such as computer vision, speech processing and natural language understanding. However, while research at the intersection of FL and LLMs is growing, it is still limited \cite{DBLP:conf/nldb/HilmkilCBSZM21, DBLP:conf/emnlp/ZhengZWQZZ24, DBLP:conf/nips/Ye0ZCDL0C24, DBLP:journals/tmis/LiuZZGZWQ25}, and studies specifically addressing multilingual FL with LLMs are even scarcer \cite{DBLP:conf/naacl/WellerMBLD22, DBLP:conf/naacl/GuoZZXK24, DBLP:journals/corr/abs-2502-04387, DBLP:journals/corr/abs-2507-03003}, as the setting introduces additional difficulties due to language heterogeneity and highly imbalanced resource distributions across languages. 

The authors in \citet{DBLP:conf/naacl/GuoZZXK24}
propose a federated multilingual framework for language modeling and text classification tasks that integrates LoRA-based parameter-efficient tuning with language-family clustering, thereby reducing communication overhead while alleviating cross-lingual interference across heterogeneous language distributions. The work in \citet{DBLP:journals/corr/abs-2502-04387} introduces an FL approach that enables clients to collaboratively learn personalized PEFT configurations, such as LoRA ranks. The method automatically discovers language-agnostic rank structures and facilitates cross-lingual transfer. 
According to \citet{DBLP:conf/naacl/WellerMBLD22}, fine-tuning pretrained multilingual models in FL mitigates the performance degradation typically seen in FL, yielding performance close to (and sometimes better than) centralized training, even when clients are partitioned by language. This suggests pretrained multilingual models are a strong practical choice for federated settings with diverse private language data. 

Beyond multilinguality, extensive research has addressed the effects of data heterogeneity (i.e., non-IID data distributions) in FL \cite{DBLP:journals/corr/abs-1912-04977, DBLP:conf/cvpr/MendietaYW0D022, DBLP:journals/csur/YeFDYT24}, a key factor that contributes to the well-known problem of client drift, which hinders convergence and degrades global model performance. The FL community has proposed a range of approaches to mitigate this issue, typically categorized into data-centric and model-centric strategies \cite{DBLP:journals/tnn/TanYCY23}. 

Data-centric approaches seek to reduce distributional discrepancies through data augmentation or shared proxy datasets \cite{DBLP:journals/corr/abs-1806-00582}. However, these methods often assume access to public or shared data, which weakens FL’s privacy guarantees. Other techniques, such as FAVOR \cite{DBLP:conf/infocom/WangKNL20}, address heterogeneity by selectively sampling clients in each communication round to balance non-IID effects. Model-centric strategies, in contrast, focus on modifying the training process or optimization objectives. For instance, FedProx \cite{DBLP:conf/mlsys/LiSZSTS20} introduces a proximal term to constrain local updates and stabilize convergence under heterogeneous data conditions, while SCAFFOLD \cite{DBLP:journals/corr/abs-1910-06378} reduces client drift using control variates on both the client and server sides to correct biased updates.

Despite this progress, the intersection of multilinguality, LLM fine-tuning and federated optimization dynamics still remains largely underexplored. Our work contributes to filling this gap by systematically analyzing how alternative multilingual client-level data distributions impact both convergence behavior and final task performance in federated fine-tuning of multilingual LLMs. Concretely, we test in controlled experiments whether increasing within-client multilinguality reduces client drift.

 \begin{figure}[tp]
   \centering
    \includegraphics[width=1\linewidth]{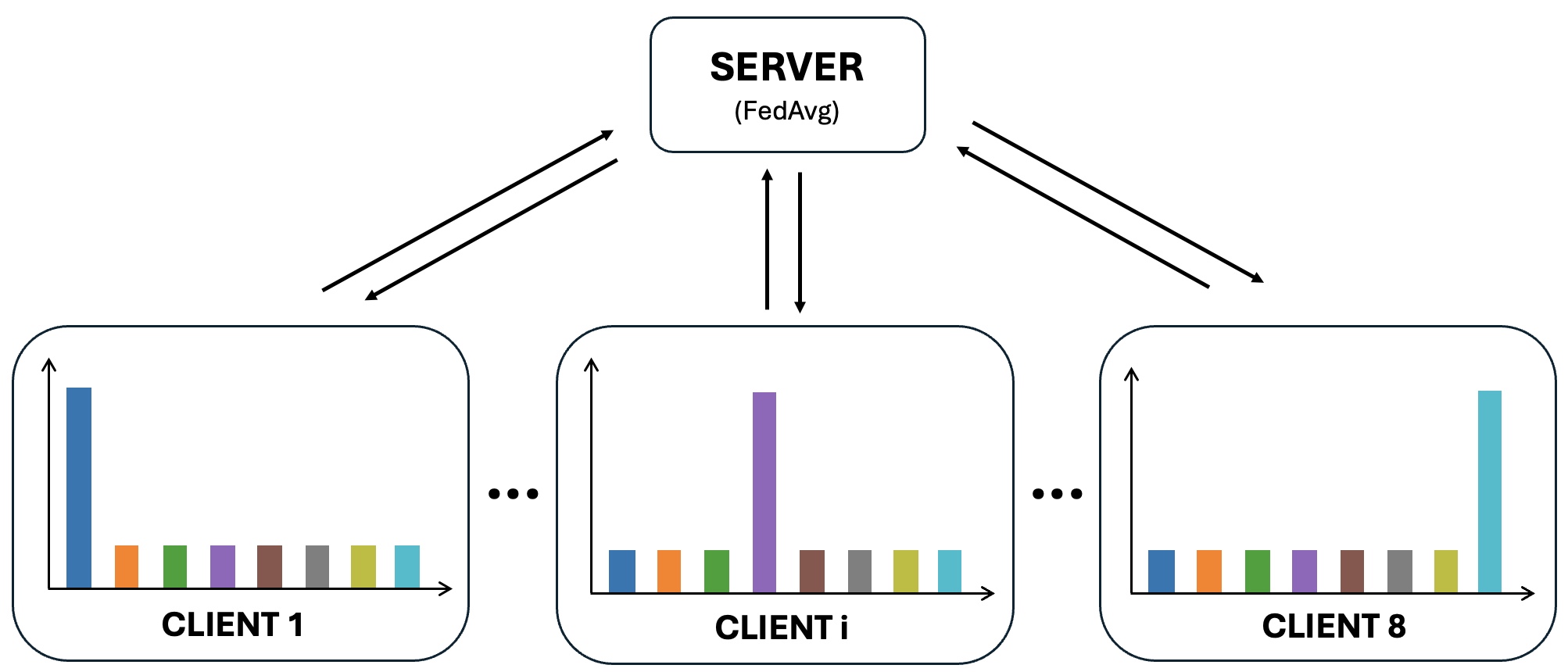}
    \vspace{-15pt}
    \caption{Illustration of a multilingual FL setup. Each client primarily contains data from a single dominant language (represented by the tallest bar) along with smaller, equal portions of data from the remaining languages. Each color corresponds to a different language. As in our experiments, there are eight clients in total.}
    \label{fig:scheme}
\end{figure}

\section{Multilingual FL Adaptation}
\label{sec:format}
In the standard FL scenario, let $\mathcal{D}_T$ denote the overall training dataset partitioned across the $|\mathcal{C}|$ clients into local subsets $\mathcal{D}_{T_i}$. The server then aggregates the weights $\mathbf{w}_i$ sent by the clients as:
\begin{equation}
\mathbf{w} = \sum_{i=1}^{|\mathcal C|}\alpha_i \mathbf{w}_i,
\label{equ:ParamAvg}
\end{equation}
where $\mathbf{w}_i$ is the parameter vector trained by client $i$, $\mathbf{w}$ is the parameter vector after aggregation on the server, $\mathcal{C}$ is the set of clients, and $\alpha_i = \frac{|\mathcal{D}_{T_i}|}{|\mathcal{D}_T|} \geq 0$ denotes the aggregation weight of client $i$, with $\sum_{i=1}^{|\mathcal C|}\alpha_i = 1$. Here, $|\mathcal{D}_{T_i}|$ is the size of the local training set of client $i$, and $|\mathcal{D}_T| = \sum_{i=1}^{|\mathcal C|} |\mathcal{D}_{T_i}|$ is the total number of training samples across all clients. Formally, the optimization problem can be expressed as:
\begin{equation}
\min_{\mathbf{w} \in \mathbb{R}^d} F(\mathbf{w}) := \frac{1}{|\mathcal C|} \sum_{\mathcal C_i}F_i(\mathbf{w})
\label{equ:Global objective function}
\end{equation}
\quad where 
\begin{equation}
F_i(\mathbf{w}):= \mathbb{E}_{(x,y) \sim \mathcal{D}_{T_i}} \left[ F_i(\mathbf{w}; x, y) \right]
\label{equ:loss}
\end{equation}
denotes the expected loss of the $i$-th client over its local dataset $\mathcal{D}_{T_i}$ (i.e., the expectation is taken over all data pairs $(x, y)$ sampled from the client’s data distribution.) In the context of instruction-tuning, $x$ represents the input (consisting of the instruction and associated prompt) and $y$ is the corresponding ground-truth response. For more details, see Section~\ref{ssec:model}. 

Aggregation strategies in FL range from the widely used FedAvg \cite{DBLP:journals/corr/McMahanMRA16} to optimization-aware methods like FedProx \cite{DBLP:journals/corr/abs-1812-06127}, SCAFFOLD \cite{DBLP:journals/corr/abs-1910-06378} or FedOpt \cite{DBLP:journals/corr/abs-2003-00295}. In our federated settings, the size of each client’s local training dataset $|\mathcal{D}_{T_i}|$ is the same for all clients. Thus, all $\alpha_i$ in equation \ref{equ:ParamAvg} have the same value, resulting in a FedAvg algorithm where all clients contribute the same to the global model.

\subsection{Parameter-Efficient Fine-Tuning}
\label{ssec:PEFT}

Parameter-Efficient Fine-Tuning (PEFT) is highly effective in fine-tuning LLMs. In PEFT, most model parameters remain frozen, and only a small targeted subset is optimized for downstream tasks \cite{houlsby2019}. Although this significantly reduces computational and memory overhead, by updating only a small subset of parameters -- typically around $1\%$ or $2\%$ of the whole pre-trained model --, empirical findings in \citet{zhang2023d} indicate that incorporating PEFT techniques can sometimes reduce language model performance.

PEFT techniques, classified by \citet{ding2022}, fall into three distinct categories. Addition-based approaches, exemplified by adapters \cite{houlsby2019}, augment the original model with new trainable elements. Specification-based approaches, such as diff pruning \cite{guo2021}, refine the fine-tuning process by selectively activating existing parameters. Lastly, reparameterization-based methods, including LoRA \cite{DBLP:conf/iclr/HuSWALWWC22}, optimize fine-tuning efficiency through reparameterization strategies.

While the formulations in \ref{equ:ParamAvg}, \ref{equ:Global objective function} and \ref{equ:loss} are written in a generic way, in the experiments conducted in this work, $w_i$ denotes the set of trainable LoRA parameters $(A_i, B_i)$ of client $i$ (i.e., the collection of LoRA matrices across all adapted layers), whereas $w$ denotes the aggregated global LoRA parameters. Concretely, given a frozen pre-trained weight matrix $W_0$, LoRA learns two low-rank matrices $A_i$ and $B_i$ for each client $i$, which induce an update $\Delta W_i = B_iA_i$. Hence, the adapted weight matrix becomes $W_0 + \Delta W_i$, and the federated communication and aggregation operate only on these trainable LoRA parameters, rather than on the full backbone parameters.

\subsection{FederatedScope Extension for multilingual FL with LLMs}

To support multilingual FL with LLMs, we extended \textit{FederatedScope-LLM} with the necessary modifications to handle multilingual scenarios, allowing each client to operate on language-specific data. We developed tools for multilingual splitting that include a shared multilingual validation and test set for the global model $\mathbf{w}$ in the server, alongside language-specific training, validation and test sets for each client. This partitioning reflects a realistic FL setting in which the server and clients are distinct entities, and each client $i$ holds a private local dataset $\mathcal{D}_i$ consisting exclusively of examples in a single language. We refer to this configuration in our experiments as 100\% mono, which means fully monolingual clients.

Building on the initial scenario, we also prepared configurations with increasing degrees of client multilinguality (85\% mono, 70\% mono, 50\% mono, etc.). These settings were created by redistributing part of each client’s monolingual data across other clients and replacing it with samples from different languages. As a result, the share of same-language data per client decreases as within-client multilinguality increases. Figure \ref{fig:scheme} illustrates this multilingual client composition. Importantly, the total number of training examples per client is kept constant in all settings. Although real-world multilingual FL deployments often involve strongly imbalanced per-language and per-client data volumes, we enforce equal per-client dataset sizes to maintain a controlled experimental setting and isolate the impact of multilinguality from confounding effects due to diverse client dataset sizes.

\section{Local Dynamic Early Stopping}
\label{subsec:stop} 
We introduce Local Dynamic Early Stopping for Federated Learning (LDES-FL), a method that enables each client to autonomously decide when to stop local training based on validation performance on its private validation dataset.
Each client monitors its validation loss, $F_{i}({\mathbf{w}}^{(t)})$, and stops training once no further improvement is observed. The client then periodically sends its best-performing local model to the server for aggregation. The server continues to perform model averaging across all clients, but for those halted clients, it employs their best local models in subsequent communication rounds. 

\begin{figure}[tp]
   \centering
    \includegraphics[width=1\linewidth]{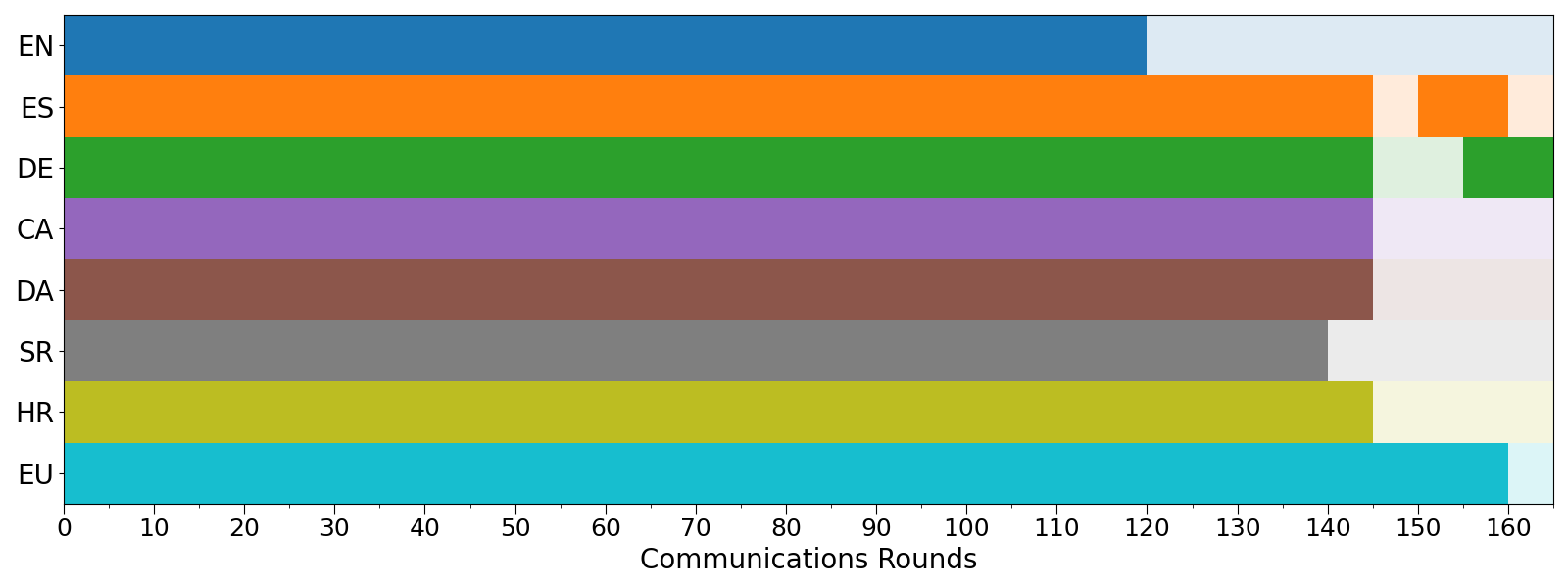}
    \vspace{-15pt}
    \caption{Training evolution of clients using LDES-FL with FedAvg, where each client holds data in a different language (100\% mono).}
    \label{fig:clients_evolution}
\end{figure}

Crucially, our algorithm supports adaptive client rejoining. Even after stopping local training, stopped clients continue to evaluate the validation loss of the downloaded global model on their local validation dataset. A client that previously stopped may resume local training if the downloaded model yields an improvement in its local validation loss, $F_i(\mathbf{w}^{(t)})$. Federated training terminates once all clients are stopped. This process is illustrated in Figure~\ref{fig:clients_evolution}, where colored regions indicate active training periods and blank regions represent idle periods after local early stopping. Note that clients \texttt{ES} and \texttt{DE} both stop and later resume training. For a better understanding of the algorithm, see Algorithm~\ref{alg:algorithm}. In the algorithm, $a_i$ indicates whether client $\mathcal{C}_i$ is active (1) or stopped (0), while $\mathbf{w}$ is used as a generic notation for the trainable client parameters. In our LoRA-based setting, these correspond to the LoRA matrices $A_i$ and $B_i$ for each client.

\begin{figure}[tp]
    \centering
    \includegraphics[width=1\linewidth]{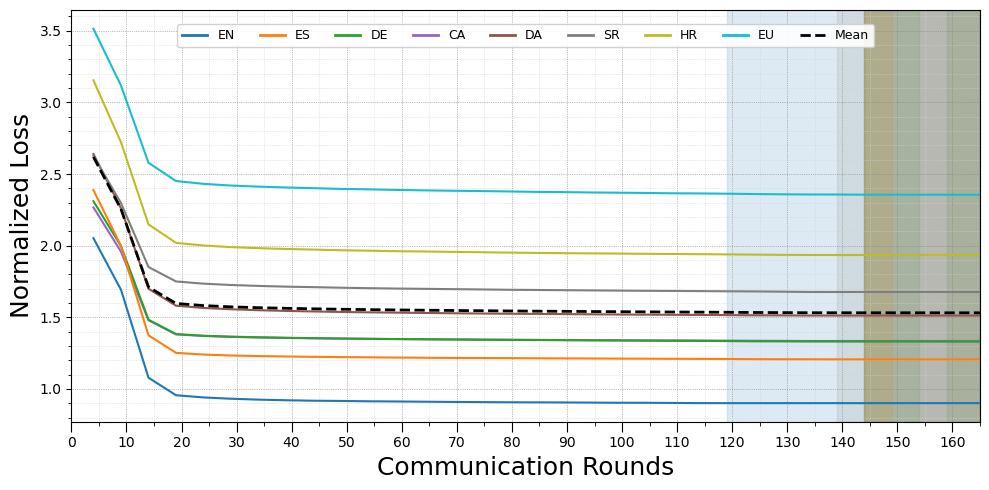}
    \vspace{-15pt} 
    \caption{Validation loss across clients under standard FedAvg (100\% mono). Low-resource languages show higher validation loss, whereas high-resource languages achieve lower loss. All clients improve at a similar rate during federated training. Shaded regions mark rounds where a client is stopped.}
    \label{fig:curves}
\end{figure}

This dynamic participation mechanism contrasts with standard federated early stopping, in which global training termination is determined by a global validation loss (e.g., the mean of all clients’ losses; black dashed curve in Figure~\ref{fig:curves}), as implemented in \textit{FederatedScope}. In contrast, LDES-FL enables personalized convergence, allowing clients to stop and resume training independently.

Figure~\ref{fig:curves} shows both the best validation loss for each monolingual client and the average across clients for the 100\% mono setting. The order of the languages in the figure reflects their resource availability in the model used (\texttt{salamandra-2b-instruct}), based on the amount of data used during pretraining. Languages with more pretraining data (high-resource languages) tend to perform better, yielding lower losses, and are shown at the bottom. In contrast, languages with less pretraining data (low-resource languages) tend to perform worse, yielding higher losses, and are shown at the top.

\section{Multilingual Experiments} 
\subsection{Data}
\label{ssec:data}
We employ a multilingual version of the \textsc{Alpaca Cleaned} dataset covering eight languages with their corresponding ISO 639-1 codes: English (\texttt{en}), Spanish (\texttt{es}), German (\texttt{de}), Catalan (\texttt{ca}), Danish (\texttt{da}), Serbian (\texttt{sr}), Croatian (\texttt{hr}) and Basque (\texttt{eu}), each containing 52{,}002 samples. The original \textsc{Alpaca} dataset, introduced by Stanford for instruction-tuning \cite{alpaca2023}, was later refined using GPT-4 to improve consistency and formatting \cite{alpaca_cleaned2023}. This cleaned version was subsequently expanded through large-scale English-to-XX translation \cite{DBLP:journals/corr/abs-2311-10797}. From this resource, we constructed data partitions specifically designed for our FL experiments, ensuring comparable data volumes and distributions across clients in the multilingual instruction-tuning setup. The datasets used in our experiments can be downloaded from Hugging Face via the instructions provided in the repository\textsuperscript{\ref{fn:repo}}.

Each federated client receives 48{,}960 training samples and 1{,}020 validation samples. As introduced earlier, in our experiments the language composition of each client’s data varies across settings, ranging from fully monolingual (i.e., clients contain data from a single language) to fully multilingual (i.e., all clients share data from all languages). In the mixed settings, clients contain different ratios of monolingual and multilingual data, which determines the level of cross-client language heterogeneity in the federation. The specific data distributions evaluated in our experiments are summarized in Table~\ref{tab:fl_scenarios}.

\begin{algorithm}[!t]
\caption{Local Dynamic Early Stopping}
\label{alg:algorithm}
\LinesNumbered
\KwData{$\mathcal D_i$, $\mathbf{w}^{(0)}$, $a_i \in \{0,1\}$, $\mathcal C_i$, $P_{\max}$}

Initialize $t \leftarrow 1$, $p_i \leftarrow 0$, $\mathbf{w}_i^{(0)} \leftarrow \mathbf{w}^{(0)}$, $\mathbf{w}_i^{\text{best}} \leftarrow \mathbf{w}^{(0)}$, and $a_i \leftarrow 1$, $\forall i \in \mathcal C$\;

\While{$\sum_i a_i > 0$}{

    \textbf{Local training process:}\ \\
    \For{$\mathcal C_i \in \mathcal C$}{
        \If{$a_i = 1$}{
            \small Update the local parameters $\mathbf{w}_i^{(t)}$ \\
            using local TRAIN data $\mathcal D_{T_i} \in \mathcal D_i$\;
            $\mathbf{w}_i^{(t)} \leftarrow \arg\min\limits_{\mathbf{w}_i} F_i(\mathbf{w}_i)$\;
        }
    }

    \textbf{Model aggregation process:}\ \\
    {\small The server computes weights}\ \\
    $\mathbf{w}^{(t)} = \sum\limits_{i=1}^{N} \alpha_i \mathbf{w}_i^{(t)}$ with
    $\alpha_i = \frac{\lvert \mathcal D_{T_i}\rvert}{\lvert \mathcal D_T\rvert}$\;
    {\small \text{The server broadcasts global parameters}} \\

    \textbf{Local validation process:}\ \\
    \For{$\mathcal C_i \in \mathcal C$}{
        \small Validate the global parameters $\mathbf{w}^{(t)}$ \\
        using local VAL data $\mathcal D_{V_i} \in \mathcal D_i$\;

        \If{$F_i(\mathbf{w}^{(t)}) > F_i(\mathbf{w}_i^{\text{best}})$}{
            $p_i \leftarrow p_i + 1$\;
            \If{$p_i \geq P_{\max}$}{
                $a_i \leftarrow 0$ stopped; \\
                $\mathbf{w}_i^{(t)} \leftarrow \mathbf{w}_i^{\text{best}}$\;
            }
        }
        \Else{
            $a_i \leftarrow 1$ resumed; \\
            $p_i \leftarrow 0$\;
            $\mathbf{w}_i^{\text{best}} \leftarrow \mathbf{w}^{(t)}$\;
        }
    }

    Clients upload local VAL loss $F_i(\mathbf{w}^{(t)})$\;
    $t \leftarrow t + 1$\;
}
\KwResult{$\mathbf{w}^{(t)}$}
\end{algorithm}

\begin{table}[!tp]
\centering
\begin{tabular}{lc}
\toprule
\textbf{Client Composition} & \textbf{Heterogeneity Level} \\
\midrule
100\% mono - 0\% multi & \hetscale{6} \\
85\% mono - 15\% multi & \hetscale{5} \\
70\% mono - 30\% multi & \hetscale{4} \\
50\% mono - 50\% multi & \hetscale{3} \\
30\% mono - 70\% multi & \hetscale{2} \\
15\% mono - 85\% multi & \hetscale{1} \\
\bottomrule
\end{tabular}
\caption{Client language compositions used to control cross-client language heterogeneity in our FL experiments. More filled circles indicate higher heterogeneity and stronger cross-client non-IIDness (i.e., larger monolingual share), whereas fewer filled circles indicate a more IID setting in which clients have more similar language distributions.}
\label{tab:fl_scenarios}
\end{table}

In all the federated settings, the central server holds no training data. Instead, it maintains a fixed multilingual test set containing 501 samples per language (4{,}008 samples in total) used exclusively for global evaluation. 

Client data was randomly partitioned without enforcing cross-lingual alignment, preserving only the desired language proportions per client. Consequently, translated instances can end up in different splits across clients (e.g., an example used for training in one language may appear in the validation or test split in another). Throughout all FL experiments, training and validation data remain client-side, while the server uses only the held-out multilingual test set for evaluation.

Although the multilingual dataset contains 52,002 instances per language, the experiments reported here rely only on the fixed partitions described above. Beyond the client-side training and validation splits and the held-out server-side test set, we also defined additional disjoint subsets that are not used in the current study. These reserved subsets are intended for auxiliary analyses and future extensions beyond
the current FL experiments, including client-side testing and server-side validation.

\subsection{Model and Prompt}
\label{ssec:model}
As the backbone model for all experiments, we employ the \texttt{salamandra-2b-instruct} model \cite{DBLP:journals/corr/abs-2502-08489}, a multilingual instruction-tuned LLM whose pre-training corpus covers 35 European languages and code, including the languages considered in our experiments. To reduce training costs while maintaining competitive performance, we fine-tune the model using LoRA adapters with rank $r=16$ and scaling factor $\alpha=32$ (see Section~\ref{ssec:PEFT}). For prompting the model, we use the popular ChatML-style template (as recommended by the model creators) plus the original prompt template format from the \textsc{Alpaca} dataset for processing samples. We translated the English Alpaca-style template into the seven additional languages used in our experiments, ensuring that each sample is processed with the prompt in its corresponding language. The German version, for example, is shown below. 
\vspace{12pt}

\begin{minipage}{\linewidth}
\footnotesize
\raggedright\ttfamily
Nachfolgend finden Sie eine Anweisung, \\
die eine Aufgabe beschreibt, gepaart \\
mit einer Eingabe, die weiteren \\
Kontext liefert. \\
Schreiben Sie eine Antwort, die die \\
Anfrage angemessen ergänzt.\\[4pt]
\#\#\# Anweisung: \{Instruction in German\}\\[2pt]
\#\#\# Eingabe: \{Input in German\} \\[2pt]
\#\#\# Antwort: \{LLM response\}
\label{tab:prompt}
\end{minipage}
\vspace{-0.1cm}
\subsection{Methodology}
\label{ssec:methodology}
First, we performed preliminary experiments comparing LDES-FL with standard federated early stopping under the same federated fine-tuning setup (FedAvg, patience $=1$) in two client language compositions: 100\% mono and 15\% mono. The results, presented in Section~\ref{sec:analysis}, motivate our use of LDES-FL in the subsequent federated experiments.

Then, we fine-tuned separate models for each of the eight languages in a centralized, monolingual setting. For each language, we initialized the LoRA adapters using the default PEFT configuration, with matrix \(A\) randomly initialized and matrix \(B\) initialized to zero, and trained the model on the corresponding training split \(\mathcal{D}_{T_i}\). Standard early stopping was applied based on validation loss, with a patience of 5 epochs and a minimum improvement threshold of 0.001. Validation loss was computed after every epoch. This setup corresponds to a purely local learning baseline, in which each client optimizes its own model without any communication or parameter sharing with other languages. These models are reported as \textit{Local FT (lang)} in Table~\ref{tab:server_results}.

We also fine-tuned a single model on the union of all client data, \(\mathcal{D} = \bigcup_i \mathcal{D}_i\) (i.e., the whole multilingual dataset), shuffling samples across languages and using the same LoRA initialization and  early-stopping configuration as in the monolingual local fine-tunings. Since this dataset is eight times larger, validation loss was computed after processing a portion of training data equivalent to a single client dataset, ensuring a comparable evaluation frequency for early stopping. Results for this setting are reported as \textit{Local FT (multilingual)} in Table~\ref{tab:server_results}.

Following this, we fine-tuned the  model in a federated setting using the FedAvg algorithm across the multilingual FL scenarios described in Section~\ref{ssec:data}, again starting from the default LoRA configuration. All federated experiments used the LDES-FL mechanism as the stopping criterion, with a local patience of 1. This parameter specifies the number of consecutive validation rounds without improvement that each client can tolerate before halting its local training. It is denoted by $P_{max}$ in Algorithm~\ref{alg:algorithm}, while $p_{i}$ represents the corresponding counter for client $i$. In all FL setups, each active client performed 160 local mini-batch steps per round with micro-batch size 2 and gradient accumulation over 16 steps, corresponding to 10 optimizer updates per round. This yields an effective local batch size of 32 samples per GPU per optimizer update and 320 processed samples per GPU per round. Optimization was performed using OneBitAdam (learning rate 0.001, gradient clipping norm 1.0) on cross-entropy loss. Validation was conducted every 5 training rounds.

\section{Analysis and Discussion}
\label{sec:analysis}
For the preliminary early-stopping experiments shown in Table~\ref{tab:ldes_metrics}, LDES-FL reduces unnecessary computation relative to standard federated early stopping while maintaining comparable performance. In the 100\% mono setting, LDES-FL preserves performance, slightly reduces cross-lingual dispersion, and lowers the number of optimization steps by approximately 22\%. In the 15\% mono setting, performance again remains essentially unchanged, while the number of optimization steps decreases by approximately 32\%. For these reasons, we adopted LDES-FL for the rest of the FL experiments.

Having established the efficiency of LDES-FL, we now turn to the main results. Table~\ref{tab:server_results} reports the performance of all models on the multilingual test set of the server. Inference was performed with greedy decoding to ensure reproducibility. \emph{Base Model} refers to the original pretrained instruction-tuned model (\texttt{salamandra-2b-instruct}) before any additional adaptation.

\begin{table}[tp]
\centering
\scriptsize
\setlength{\tabcolsep}{3pt}
\begin{tabular}{l|l|cc|cc|c}
\toprule
 &  & \multicolumn{2}{c|}{\textbf{Mean}\,\,\,$\uparrow$} & \multicolumn{2}{c|}{$\boldsymbol{\sigma}\,\downarrow$} & \multirow{2}{*}{ \shortstack[c]{\textbf{Optim.}\\[-0.25ex]\textbf{steps}}} \\
 \cmidrule(lr){3-6}
 &  & R & F\textsc{b} & R & F\textsc{b} & \\
\midrule
100\% mono
  & Standard & 0.202 & 0.877 & 6.75e-2 & 1.29e-2 & 1.44e+04 \\
  & LDES     & 0.203 & 0.877 & 6.47e-2 & 1.26e-2 & 1.12e+04 \\
\midrule
15\% mono
& Standard & 0.224 & 0.880 & 6.09e-2 & 1.22e-2 & 1.11e+05 \\
  & LDES     & 0.221 & 0.880 & 6.09e-2 & 1.23e-2 & 7.57e+04 \\
\bottomrule
\end{tabular}
\caption{Performance and optimization steps obtained with two early stopping methods (standard federated early stopping and LDES-FL) under two client language compositions. Federated fine-tuned models are evaluated on the multilingual test set described in Section~\ref{ssec:data}. R and F\textsc{b} denote ROUGE-L and F\textsc{bert}, respectively.}
\label{tab:ldes_metrics}
\end{table}

\begin{table*}[tp]
\centering
\resizebox{\textwidth}{!}{
\begin{tabular}{l | *{8}{cc} | cc | cc}
\toprule
 & \multicolumn{2}{c}{\textbf{EN}} & \multicolumn{2}{c}{\textbf{ES}} & \multicolumn{2}{c}{\textbf{DE}} & \multicolumn{2}{c}{\textbf{CA}} & \multicolumn{2}{c}{\textbf{DA}} & \multicolumn{2}{c}{\textbf{SR}} & \multicolumn{2}{c}{\textbf{HR}} & \multicolumn{2}{c}{\textbf{EU}} & \multicolumn{2}{|c|}{\textbf{Mean}\,\,\,$\uparrow$} & \multicolumn{2}{c}{$\boldsymbol{\sigma}\,\downarrow$} \\
\cmidrule(lr){2-3} \cmidrule(lr){4-5} \cmidrule(lr){6-7} \cmidrule(lr){8-9}
\cmidrule(lr){10-11} \cmidrule(lr){12-13} \cmidrule(lr){14-15} \cmidrule(lr){16-17} \cmidrule(l){18-19} \cmidrule(l){20-21}
 & R & F\textsc{b} & R & F\textsc{b} & R & F\textsc{b} & R & F\textsc{b} & R & F\textsc{b} & R & F\textsc{b} & R & F\textsc{b} & R & F\textsc{b} & R & F\textsc{b} & R & F\textsc{b} \\
\midrule
Base Model
& 0.307 & 0.893 & 0.215 & 0.877 & 0.170 & 0.873 & 0.207 & 0.871 & 0.160 & 0.869 & 0.074 & 0.861 & 0.126 & 0.856 & 0.080 & 0.840 & 0.167 & 0.867 & 7.70e-2 & 1.57e-2 \\
\midrule
Local FT (EN)
& 0.351 & 0.903 & 0.232 & 0.882 & 0.187 & 0.876 & 0.233 & 0.878 & 0.186 & 0.875 & 0.131 & 0.869 & 0.147 & 0.863 & 0.105 & 0.852 & 0.197 & 0.875 & 7.73e-2 & 1.49e-2 \\
Local FT (ES)
& 0.295 & 0.897 & 0.248 & 0.883 & 0.188 & 0.877 & 0.196 & 0.876 & 0.187 & 0.875 & 0.168 & 0.868 & 0.159 & 0.866 & 0.102 & 0.850 & 0.193 & 0.874 & 5.80e-2 & 1.47e-2 \\
Local FT (DE)
& 0.322 & 0.898 & 0.237 & 0.881 & 0.212 & 0.883 & 0.233 & 0.880 & 0.189 & 0.877 & 0.181 & 0.871 & 0.165 & 0.870 & 0.105 & 0.851 & 0.206 & 0.876 & 6.32e-2 & 1.34e-2 \\
Local FT (CA)
& 0.296 & 0.893 & 0.131 & 0.866 & 0.168 & 0.871 & 0.253 & 0.884 & 0.146 & 0.865 & 0.145 & 0.860 & 0.104 & 0.849 & 0.065 & 0.834 & 0.163 & 0.865 & 7.61e-2 & 1.86e-2 \\
Local FT (DA)
& 0.315 & 0.895 & 0.241 & 0.882 & 0.185 & 0.875 & 0.244 & 0.881 & 0.228 & 0.886 & 0.168 & 0.868 & 0.151 & 0.863 & 0.109 & 0.849 & 0.205 & 0.875 & 6.47e-2 & 1.45e-2 \\
Local FT (SR)
& 0.259 & 0.883 & 0.184 & 0.862 & 0.154 & 0.862 & 0.174 & 0.860 & 0.107 & 0.848 & \textbf{0.207} & 0.876 & 0.024 & 0.869 & 0.046 & 0.819 & 0.144 & 0.860 & 8.03e-2 & 1.97e-2 \\
Local FT (HR)
& 0.263 & 0.887 & 0.229 & 0.879 & 0.190 & 0.877 & 0.207 & 0.874 & 0.181 & 0.874 & 0.028 & 0.866 & 0.211 & \textbf{0.881} & 0.099 & 0.844 & 0.176 & 0.873 & 7.62e-2 & 1.31e-2 \\
Local FT (EU)
& 0.273 & 0.888 & 0.224 & 0.877 & 0.182 & 0.871 & 0.223 & 0.876 & 0.178 & 0.872 & 0.163 & 0.862 & 0.145 & 0.859 & 0.143 & 0.864 & 0.191 & 0.871 & 4.52e-2 & \textbf{9.40e-3} \\
\midrule
Local FT (multilingual)
& \textbf{0.353} & \textbf{0.905} & \textbf{0.258} & \textbf{0.886} & \textbf{0.223} & \textbf{0.885} & \textbf{0.260} & \textbf{0.886} & \textbf{0.231} & \textbf{0.887} & 0.203 & \textbf{0.878} & \textbf{0.217} & 0.880 & \textbf{0.147} & \textbf{0.865} & \textbf{0.237} & \textbf{0.884} & \textbf{5.90e-2} & 1.12e-2 \\
\midrule
FedAvg (100\% mono)
& 0.330* & 0.900* & 0.229* & 0.881* & 0.194* & 0.878* & 0.237* & 0.880* & 0.189* & 0.878* & 0.162* & 0.871* & 0.174* & 0.871* & 0.110* & 0.854* & 0.203 & 0.877 & 6.47e-2 & 1.29e-2 \\
FedAvg (85\% mono)
& 0.334* & 0.900* & 0.229* & 0.881* & 0.197* & 0.879* & 0.242* & 0.881* & 0.193* & 0.878* & 0.157* & 0.872* & 0.173* & 0.872* & 0.112* & 0.854* & 0.205 & 0.877 & 6.64e-2 & 1.28e-2 \\
FedAvg (70\% mono)
& 0.335* & 0.900* & 0.232* & 0.881* & 0.206* & 0.880* & 0.240* & 0.881* & 0.203* & 0.880* & 0.154* & 0.872* & 0.181* & 0.872* & 0.114* & 0.855* & 0.208 & 0.878 & 6.57e-2 & 1.28e-2 \\
FedAvg (50\% mono)
& 0.339* & 0.901* & 0.238* & 0.882* & 0.210* & 0.881* & 0.243* & 0.881* & 0.211* & 0.883* & 0.179* & 0.873* & 0.195* & 0.874* & 0.121* & 0.857* & 0.217 & 0.879 & 6.23e-2 & 1.23e-2 \\
FedAvg (30\% mono)
& 0.337* & 0.901* & 0.244* & 0.882* & 0.206* & 0.881* & 0.251* & 0.883* & 0.208* & 0.882* & 0.194* & 0.874* & 0.197* & 0.875* & 0.125* & 0.858* & 0.220 & 0.880 & 6.06e-2 & 1.21e-2 \\
FedAvg (15\% mono)
& 0.341* & 0.902* & 0.242* & 0.882* & 0.206* & 0.881* & 0.247* & 0.883* & 0.211* & 0.883* & 0.194* & 0.873* & 0.197* & 0.875* & 0.127* & 0.858* & 0.221 & 0.880 & 6.09e-2 & 1.22e-2 \\
\bottomrule
\end{tabular}
}
\caption{\small Results for the multilingual test set of the server. R and F\textsc{b} stand for ROUGE-L \cite{lin-2004-rouge} and F\textsc{bert} \cite{DBLP:conf/iclr/ZhangKWWA20} respectively. The language in columns corresponds to the monolingual part of the test set. Mean reports the average score across all languages, whereas $\sigma$ denotes the corresponding standard deviation, used here as an indicator of multilingual fairness. A bootstrap test with 100 sets (p-value $<$ 0.05) confirms that all federated models achieve statistically significant improvements over the \textit{Base Model} on all evaluated metrics and languages, as indicated by an asterisk (*).}
\label{tab:server_results}
\end{table*}

\begin{table}[!tp]
\centering
\small
\begin{tabular}{lc}
\toprule
\textbf{Training Setting} & \textbf{Norm. Optim. Steps} \\
\midrule
Local FT (EN) & 0.106 \\
Local FT (ES) & 0.312 \\
Local FT (DE) & 0.345 \\
Local FT (CA) & 0.362 \\
Local FT (DA) & 0.351 \\
Local FT (SR) & 0.429 \\
Local FT (HR) & 0.340 \\
Local FT (EU) & 0.385 \\
\midrule
Local FT (multilingual) & 1.000 \\
\midrule
FedAvg (100\% mono) & 0.061 \\
FedAvg (85\% mono) & 0.103 \\
FedAvg (70\% mono) & 0.146 \\
FedAvg (50\% mono) & 0.274 \\
FedAvg (30\% mono) & 0.370 \\
FedAvg (15\% mono) & 0.414 \\
\bottomrule
\end{tabular}
\caption{Training cost normalized by the highest-step configuration, \textit{Local FT (multilingual)}, which requires \(1.83\times 10^{5}\) total optimization steps. Each value is reported as a fraction of this reference.}
\label{tab:training_steps}
\end{table}

\begin{table*}[!tp]
\centering
\scriptsize
\setlength{\tabcolsep}{3.2pt}
\renewcommand{\arraystretch}{0.97}
\begin{tabular}{l|c|c|c|c|c|c|c|c|c|c|c|c}
\toprule
& \multicolumn{6}{c|}{\textbf{ROUGE-L}} & \multicolumn{6}{c}{\textbf{F\textsc{bert}}} \\
\cmidrule(lr){2-7} \cmidrule(lr){8-13}
& \textbf{Base} & \textbf{Avg. Local} & \textbf{Local FT} & \textbf{FedAvg} & \textbf{FedAvg} & \multirow{2}{*}{\textbf{\normalsize $\Delta$}}
& \textbf{Base} & \textbf{Avg. Local} & \textbf{Local FT} & \textbf{FedAvg} & \textbf{FedAvg} & \multirow{2}{*}{\textbf{\normalsize $\Delta$}} \\
& \textbf{Model} & \textbf{FT (mono)} & \textbf{(multi)} & \textbf{100\% mono} & \textbf{15\% mono} &
& \textbf{Model} & \textbf{FT (mono)} & \textbf{(multi)} & \textbf{100\% mono} & \textbf{15\% mono} & \\
\midrule
H & 0.231 & 0.232 & 0.278 & 0.251 & 0.263 & +0.012 & 0.881 & 0.881 & 0.892 & 0.886 & 0.888 & +0.002 \\
M  & 0.184 & 0.198 & 0.246 & 0.213 & 0.229 & +0.016 & 0.870 & 0.874 & 0.882 & 0.879 & 0.883 & +0.004 \\
L  & 0.093 & 0.128 & 0.189 & 0.149 & 0.173 & +0.024 & 0.852 & 0.859 & 0.868 & 0.865 & 0.869 & +0.004 \\
\bottomrule
\end{tabular}
\caption{Average performance by resource group. H, M and L stand for high-resource, mid-resource and low-resource respectively. Avg. Local FT (mono) is the average of the eight monolingual Local FT models within each resource group. 
$\Delta$ denotes the absolute improvement from 100\% mono to 15\% mono.}
\label{tab:resource_split_fedavg}
\end{table*}

A first observation is that \emph{Local FT (multilingual)} provides the strongest overall results in the table. It achieves the best aggregate performance, with a mean ROUGE-L of 0.237 and a mean F\textsc{bert} of 0.884, and it also obtains the best score in almost all per-language metrics, with only minor exceptions (e.g., SR in ROUGE-L and HR in F\textsc{bert}). This indicates that, when all multilingual data can be pooled and jointly optimized, centralized multilingual fine-tuning yields the strongest multilingual model in terms of average quality. Since this setting assumes centralized access to all training data, it should be interpreted as an upper bound for the federated setting, rather than as a directly comparable privacy-preserving alternative.
Regarding multilingual fairness, \emph{Local FT (multilingual)} also shows low standard deviations across languages ($\sigma_R = 5.90\times10^{-2}$ and $\sigma_{F\textsc{bert}} = 1.12\times10^{-2}$), consistent with its strong multilingual behavior. Although the lowest standard deviation is obtained by \emph{Local FT (EU)}, this is somewhat misleading, since that model does not achieve the best multilingual performance overall and instead produces more uniformly modest scores across languages. Thus, \emph{Local FT (multilingual)} provides a more meaningful reference point for balanced multilingual performance, since it combines high average multilingual quality with relatively low cross-lingual dispersion.

Interestingly, in most cases, the multilingual Local FT model outperforms the set of monolingual Local FT models across all languages and metrics -- even outperforming each monolingual model on the very language it was trained for. This suggests that multilingual joint optimization provides benefits beyond language-specific specialization. Such improvements may reflect positive cross-lingual transfer, whereby representations learned from multiple languages also help improve performance on individual target languages.

By contrast, monolingual Local FT models exhibit a different pattern. Each model achieves its highest performance on the language it was specifically trained on, compared to its performance on other languages. This confirms that local adaptation is highly effective for specializing the model to a single client language. However, these gains are not uniformly transferred to the rest of the multilingual test set. As a result, although monolingual Local FT improves target language performance, it generally leads to more uneven cross-lingual behavior than \emph{Local FT (multilingual)}, as reflected in its higher standard deviation and lower average multilingual performance.

Compared with the \emph{Base Model}, monolingual Local FT models often show small gains on languages that were not seen during fine-tuning. While this could suggest some degree of cross-lingual transfer, such improvements should be interpreted with caution. Since the multilingual splits are not strictly parallel, examples from different languages may still share content, structure or task-specific answer patterns. Therefore, part of the apparent transfer may arise from indirect overlap in instance format or solution templates rather than from robust language-agnostic generalization. This is an important limitation of the dataset and should be kept in mind when interpreting zero-shot transfer across languages.

On the other hand, all federated models consistently improve over the \emph{Base Model} on every reported aggregate metric. The strongest federated setting reaches a mean ROUGE-L of 0.221 and a mean F\textsc{bert} of 0.880, compared with 0.167 and 0.867 for the \emph{Base Model}. This corresponds to a substantial gain in average multilingual performance, especially in ROUGE-L. Moreover, all federated settings improve multilingual fairness, indicating that the gains are not concentrated in only a few languages, but are instead distributed more evenly across all languages. Under this criterion, FL not only improves quality, but also yields a fairer multilingual model.

Restricting the comparison to monolingual Local FT and federated adaptation, we observe an important trade‑off. Monolingual Local FT yields the strongest language-specific specialization, but at the cost of weaker multilingual balance. Federated training, by contrast, achieves less extreme specialization on any single language while providing better overall multilingual coverage. Notably, from the 50\% mono setting onward, federated models outperform all monolingual Local FT variants in mean performance, showing that aggregating updates from linguistically diverse clients produces a stronger and more balanced single model across languages. However, federated training remains below centralized multilingual fine-tuning, reflecting the cost of decentralization under privacy constraints.

The language composition of clients also has a clear effect on the behavior of FedAvg. As the proportion of multilingual data within each client increases (i.e., moving from 100\% mono to 15\% mono), average multilingual performance generally improves and the standard deviation across languages generally decreases. The best average performance is obtained in the 15\% mono setting (mean ROUGE-L 0.221, mean F\textsc{bert} 0.880), while the best fairness is observed in the 30\% mono setting ($\sigma_R = 6.06\times10^{-2}$ and $\sigma_{F\textsc{bert}} = 1.21\times10^{-2}$). This suggests that making clients more multilingual mitigates the tension between language-specific optimization and global multilingual generalization. A plausible explanation is that increasing client multilinguality reduces client-drift during federated optimization, since the local objective functions become closer approximations of the global objective, leading to more aligned gradient updates and less conflict across clients.

These improvements, however, come with a non-negligible computational cost. Table~\ref{tab:training_steps} shows that the total number of optimization steps increases sharply as clients become more multilingual. The most expensive configuration is \emph{Local FT (multilingual)}, which serves as the highest-cost upper bound. By comparison, FedAvg (100\% mono) requires only \(1.12\times10^4\) total steps, while FedAvg (15\% mono) requires \(7.57\times10^4\) steps, approximately 6.8 times more than the monolingual federated setting but still substantially less than centralized multilingual fine-tuning. In other words, strongly monolingual clients exhibit faster apparent convergence in terms of optimization steps, but much of this speed reflects early saturation at a biased local optimum induced by non-IID data, whereas more multilingual clients support longer, more productive optimization toward a better global solution. This behavior is consistent with federated optimization theory: as clients become more multilingual, their local objectives align better with the global objective, which reduces client drift and leads to higher final performance, but also prolongs training because the model continues to improve for more communication rounds instead of plateauing early at a suboptimal solution. From a practical perspective, the results suggest a trade-off frontier: highly multilingual clients yield the strongest and fairest models at the cost of substantially more optimization steps, whereas more monolingual clients converge quickly but to a less optimal and less balanced global solution. Thus, the choice depends on whether deployment prioritizes training efficiency (fewer rounds to an early plateau) or balanced multilingual quality (more rounds to a better optimum).

\begin{figure}[!tp]
   \centering
    \includegraphics[width=1\linewidth]{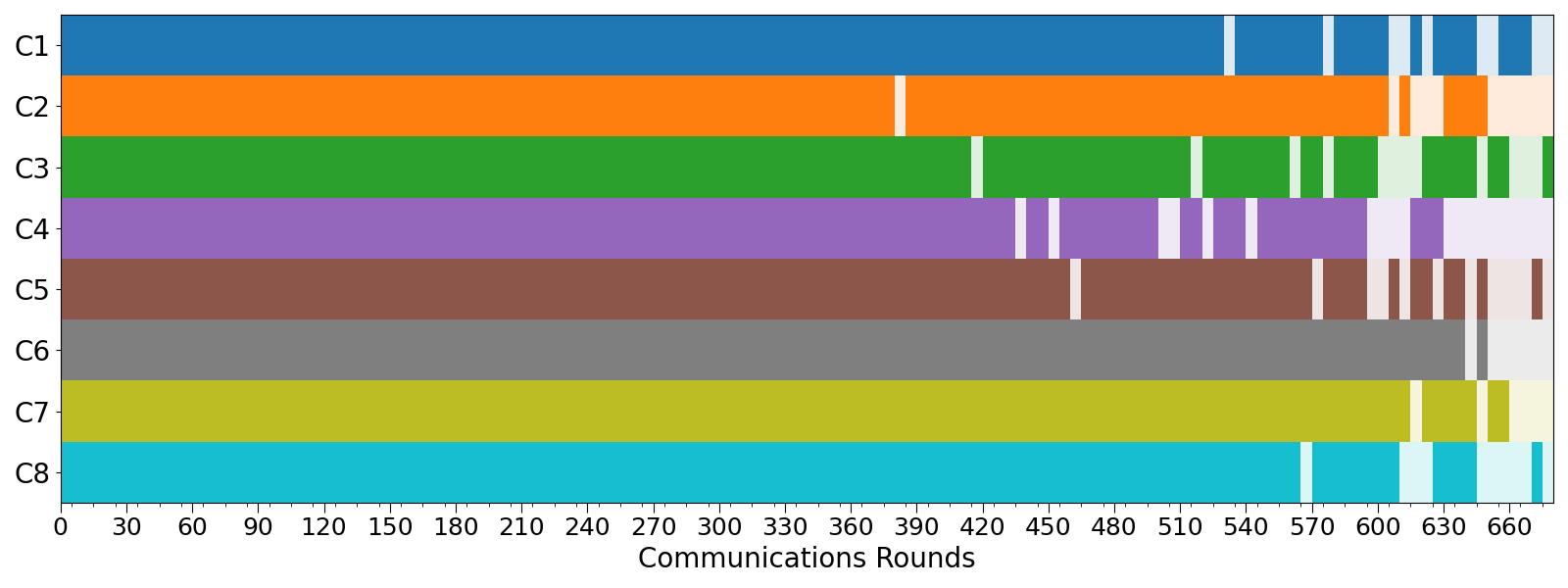}
    \vspace{-15pt}
    \caption{Training evolution of clients using LDES-FL with FedAvg in the 50\% mono setting. Note that clients are not labeled by language as in Figure \ref{fig:clients_evolution}, as here each client contains a mix of languages.}
    \label{fig:clients_evolution_mix70}
\end{figure}

As shown in Table~\ref{tab:resource_split_fedavg}, the effect of increasing client multilinguality is not uniform across resource language levels. When moving from FedAvg (100\% mono) to FedAvg (15\% mono), the average ROUGE-L score increases from 0.251 to 0.263 (+4.78\%) for the high-resource group (EN, ES, DE), from 0.213 to 0.229 (+7.51\%) for the mid-resource group (CA, DA), and from 0.149 to 0.173 (+16.11\%) for the low-resource group (SR, HR, EU). Thus, the absolute gain ($\Delta$) grows as resource availability decreases (0.012 \textrightarrow{} 0.016 \textrightarrow{} 0.024). A similar but smaller pattern is observed for F\textsc{bert}. Overall, these results suggest that making clients more multilingual is especially beneficial for lower-resource languages, and helps reduce performance disparities across language groups. 

A practical advantage of federated training is that it avoids the need for each participant to fine-tune and maintain its own separate language-specific model. Under our Local FT setup, which uses early stopping with patience 5 and $\delta=0.001$, training all eight monolingual Local FT models would require a total of $4.814\times10^5$ optimization steps, whereas the federated configurations require only between $1.12\times10^4$ and $7.57\times10^4$ steps. While this is not a perfectly controlled comparison, since Local FT and federated training do not share exactly the same stopping conditions (standard early stop vs LDES-FL), it nevertheless provides a useful estimate of relative training effort. Even the most expensive federated setting is still about 6.4 times cheaper in total training steps than the full set of Local FTs, while the cheapest one is about 43 times cheaper. Furthermore, Table~\ref{tab:resource_split_fedavg} shows that, in every resource group, the global federated model (either FedAvg composition) outperforms the \emph{average} of the monolingual Local FT models. For ROUGE-L, \emph{FedAvg (100\% mono)} exceeds \emph{Avg. Local FT (mono)} by +0.019 (H), +0.015 (M), and +0.021 (L), while \emph{FedAvg (15\% mono)} increases this margin to +0.031 (H), +0.031 (M), and +0.045 (L). The same trend holds for F\textsc{bert}. In practical terms, this means that clients can join a collaborative training workflow and obtain a single multilingual model with strong overall performance at a much lower total computational cost than training separate models in isolation. The trade-off is reduced language-specific specialization, but the resulting model is considerably more attractive when multilingual coverage is the primary goal.

Regarding the FL training scheme, the evolution of client participation confirms the proper functioning of our LDES-FL method. A comparison between Figures \ref{fig:clients_evolution} and \ref{fig:clients_evolution_mix70} shows that client rejoining occurs more frequently in settings where clients hold multilingual data, while it is almost absent in fully monolingual configurations. This behavior aligns with our expectations. In the monolingual client scenario, where both the training and validation data within each client correspond to a single language, local models learn mostly from their own data, yielding limited benefits from cross-client updates. Conversely, as the proportion of multilingual data increases, rejoining events become more frequent, since clients benefit not only from their own multilingual datasets but also from the aggregated updates contributed by other multilingual clients.

\section{Conclusions}
\label{sec:conclusions}
In this work, we extended \textit{FederatedScope}-LLM to support multilingual federated instruction-tuning of LLMs and introduced Local Dynamic Early Stopping (LDES-FL), a novel stopping criterion that allows clients to pause and resume local training based on their validation loss. This mechanism preserves performance while reducing unnecessary computation, thereby improving training efficiency and sustainability in terms of total optimization steps.

Across all studied client language compositions, ranging from fully monolingual to increasingly multilingual clients, LoRA-based federated fine-tuning consistently improves over the base model, increasing average multilingual performance and fairness. Compared to monolingual local fine-tuning, federated training yields a stronger single multilingual model, while monolingual local fine-tuning remains most effective when the goal is to maximize performance in one target language. At the same time, multilingual local fine-tuning achieves the strongest overall results, and can therefore be interpreted as an upper bound for the federated setting, since it assumes direct access to all multilingual training data without privacy restrictions.

Finally, our results indicate that client language composition is a key design variable in multilingual FL, with direct consequences for performance, fairness and efficiency. More multilingual clients generally improve average multilingual performance and fairness, and particularly benefit lower-resource languages, although at a higher training cost. This finding is consistent with the FL literature on client drift and non-IID data, and is here observed in multilingual federated instruction-tuning of LLMs. Future work should use more strictly controlled multilingual splits to better isolate true cross-lingual transfer.

\section{Acknowledgements}
Funded by the European Union's Horizon 2020. Views and opinions expressed are, however, those of the author(s) only and do not necessarily reflect those of the European Union or European Commission-EU. Neither the European Union nor the granting authority can be held responsible for them.

\section{Ethical Statement}
This study uses a multilingual dataset derived from the Alpaca Cleaned corpus, distributed under the CC BY-NC license for academic and research use. No personally identifiable information is present in the dataset and it was anonymized before release. We acknowledge that the multilingual dataset may carry linguistic and cultural biases originating from the source data and automatic translation process.

\section{Bibliographical References}\label{sec:reference}
\vspace{-1.8em}
\bibliographystyle{lrec2026-natbib}
\bibliography{lrec2026-example}

\bibliographystylelanguageresource{lrec2026-natbib}

\end{document}